\documentclass[numbers]{article}

\usepackage[dblblindworkshop, final]{neurips_2026}
\workshoptitle{BeNTO: Beyond Next Token Prediction}

\usepackage[utf8]{inputenc} 
\usepackage[T1]{fontenc}    
\usepackage{hyperref}       
\usepackage{url}            
\usepackage{booktabs}       
\usepackage{amsfonts}       
\usepackage{nicefrac}       
\usepackage{microtype}      
\usepackage{xcolor}         
\usepackage{amsmath}
\usepackage{amsthm}
\usepackage{amssymb}
\usepackage{graphicx}
\graphicspath{{../}{../images/}{./}{images/}}
\usepackage{subcaption}
\usepackage{booktabs}
\usepackage{algorithm}
\usepackage{algpseudocode}
\usepackage{float}
\usepackage{enumitem}
\usepackage{wrapfig}
\usepackage{relsize}
\usepackage{bbm}
\usepackage{tikz}
\usepackage{multirow}
\usepackage{array}
\usetikzlibrary{
  arrows.meta,
  positioning,
  calc,
  decorations.pathreplacing,
  backgrounds,
  fit
}

\title{Your Discrete Graph Diffusion Model is Secretly Memorizing}

\author{%
  Adarsh Jamadandi, Nicolas Keriven \\
     CNRS, IRISA, Rennes, France\\
\texttt{adarsh.jamadandi@irisa.fr}, \texttt{nicolas.keriven@cnrs.fr}
}

\begin{document}
\maketitle
\begin{abstract}
Discrete graph diffusion models (GDMs) corrupt the graph structure by adding noise independently to nodes and edges under a Markovian noise model. In the unconditional setting without node features, an equivariant denoiser trained on a single graph is able to generate novel graphs and sample-level metrics such as edge overlap and WL kernel similarity find no evidence of memorization. In this work, we challenge this view and show for the first time that graph diffusion models indeed exhibit memorization. Analyzing
the reverse diffusion trajectories of GDMs trained on two-block stochastic block models (SBMs), we find a strong train-data bias: the model recovers a corrupted training graph almost perfectly compared to a held-out test graph. We term this \emph{conditional memorization}, as it surfaces only when the reverse trajectory is conditioned on a corrupted training graph and is not apparent during inference. It coincides with a growing generalization gap, where the training loss improves as the loss on held-out graphs degrades, and it cannot be mitigated by early stopping. The onset of memorization always precedes the emergence of community structure in the generated graphs, thus every checkpoint earlier produces structureless graphs.  This behavior holds across all node sizes we test.
\end{abstract}

\section{Introduction}
Diffusion models \citep{pmlr-v37-sohl-dickstein15, song2020generativemodelingestimatinggradients,
ddpm, pmlr-v139-nichol21a} have recently emerged as a powerful class of generative models, with
major improvements in image \citep{zhang2023surveydiffusionbasedimage} and video generation
\citep{melnik2024videodiffusionmodelssurvey}. Motivated by this success, diffusion-based methods
have been extended to graph generation. The two most popular approaches are: embedding nodes into a continuous latent space
and applying Gaussian diffusion
\citep{pmlr-v108-niu20a, pmlr-v162-jo22a, evdaimon2024neuralgraphgeneratorfeatureconditioned},
or operating directly in the graph space through a discrete corruption process on edges
\citep{d3pm}. The latter approach has been shown to be more appropriate for generating realistic graph
structures \citep{haefeli2023diffusionmodelsgraphsbenefit,
vignac2023digressdiscretedenoisingdiffusion}, leading to a growing body of work on both
graph diffusion and discrete flow matching
\citep{gat2024discrete, qinmadeira2024defog, roos2026categoricalflowmaps}.

An interesting empirical observation made in
\citep{chen2023efficientdegreeguidedgraphgeneration,li2024graphmaker} is that we can train graph diffusion models on a single graph and still generate novel and diverse graphs without memorizing the training data. This seems counter-intuitive compared to diffusion models applied to images \citep{kadkhodaie2024generalization,bonnaire2025why} which show dataset size as one of the factors driving generalization. In contrast, the study of memorization in graph diffusion models has received little attention; the question is instead studied under the guise of model expressivity. Recent works such as
\citep{laabid2025equivariant,lawrence2025improvingequivariantnetworksprobabilistic}
show that equivariant denoisers cannot copy their training graphs exactly, as this
would require breaking the permutation symmetry of the input graphs, which equivariant models by design cannot do. A single training graph has $\Omega(n!)$ graphs in its isomorphism class \citep{yan2024swingnnrethinkingpermutationinvariance}, and an
equivariant denoiser cannot reproduce a specific graph and instead produces an average over that class. Further, the sample-level metrics commonly used to assess whether the model has reproduced the training graph such as edge overlap
\citep{edgedependency,powerindependent}, which measures the intersection of exact edge pairs between the training and generated graph, and WL kernel similarity
\citep{JMLR:v12:shervashidze11a}, which certifies that two graphs are not isomorphic, both assert no memorization. 

In this work, we challenge this view and show graph diffusion models do exhibit a form
of memorization, which we term \emph{conditional memorization}: the training graph is
recoverable from the reverse dynamics, but only when the trajectory is seeded with a
corrupted version of that graph, and is otherwise undetectable during inference. Our
contributions can be summarized as follows:
\begin{enumerate}
  \item We study the learning dynamics of single-graph diffusion models on two-block
  SBMs at $n = \{64, 128, 256, 512\}$ and show the models exhibit \emph{conditional
  memorization}, where the model restores a corrupted training graph almost perfectly
  against a held-out graph, a bias that is invisible in ordinary unconditional
  generation and surfaces only by probing the reverse diffusion trajectories.

  \item We show the minimum held-out loss is a poor proxy for model selection, since
  memorization onset ($\tau_{\texttt{mem}}$) consistently precedes the emergence of
  community structure ($\tau_{\texttt{struct}}$) across all four sizes: the model
  cannot generate the right structure without first entering the memorization regime,
  so early stopping cannot separate the two.
\end{enumerate}
\begin{figure}[h]
    \centering
    \includegraphics[width=0.55\linewidth]{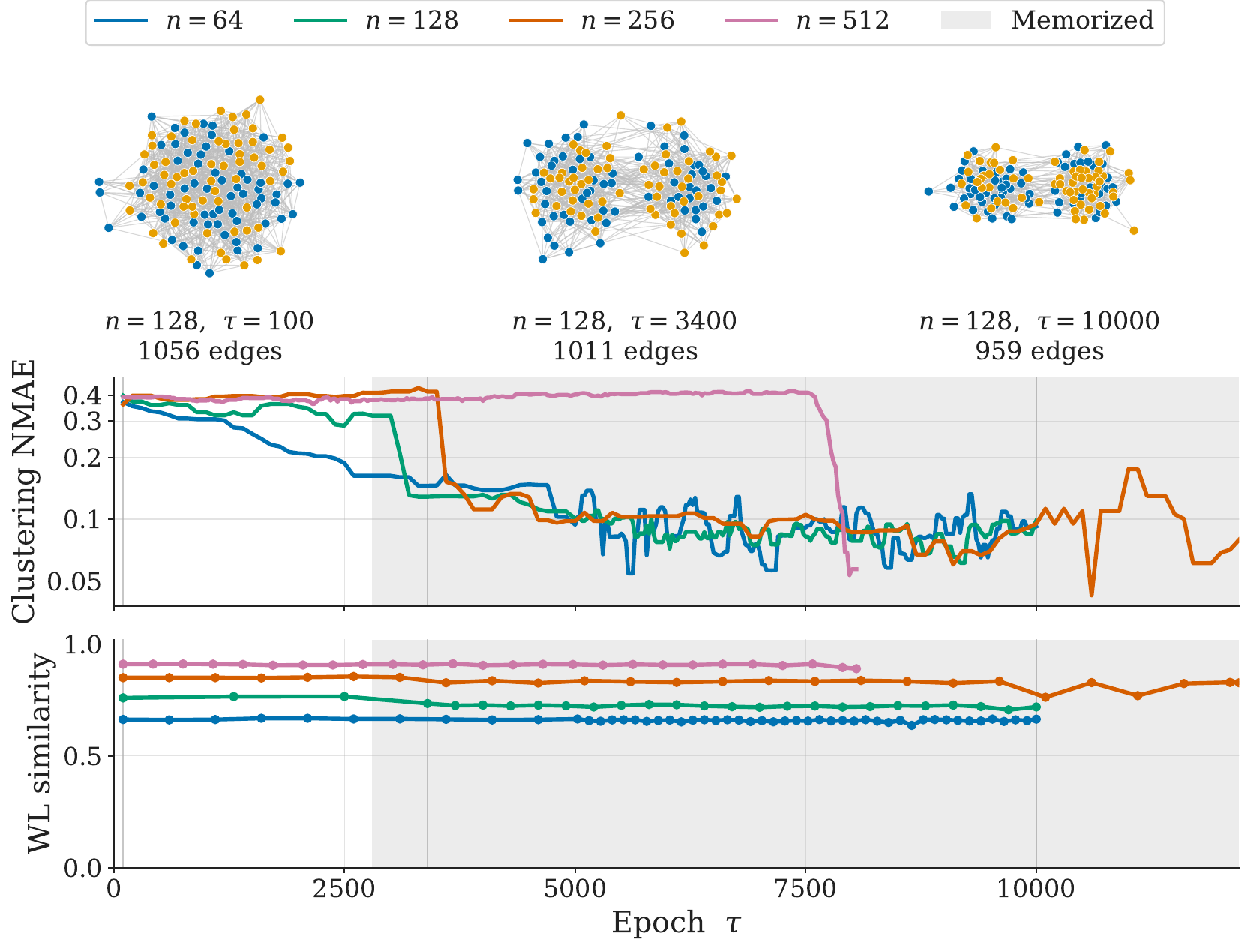}
    \caption{Sample-level metrics report a well-generalizing model. Clustering coefficient NMAE (lower is better) improves throughout training while WL kernel similarity remains flat, suggesting the samples acquire the correct community structure without reproducing the training graph.}
    \label{fig:blindmetrics}
\end{figure}

\section{Sample-level metrics are blind to memorization}
\paragraph{Data.} We generate synthetic graphs with nodes $n=\{64,128,256,512\}$ drawn from a symmetric two block stochastic block model (SBM) with block probabilities $(p_{in}, p_{out})$ chosen at each size to hold the Kesten–Stigum detectability \citep{JMLR:v18:16-480} margin fixed at 7.0 and mean degree fixed at $\approx 16$ (both well above the detectability threshold of 1), so that graph size is varied independently of task difficulty, details are present in Table \ref{tab:sbmparams}. We choose this setting as it allows us to control the true generating process and it is easier to see the \emph{structure} emerging during training.

\paragraph{Model setup.} We use EDGE \citep{chen2023efficientdegreeguidedgraphgeneration} as our backbone diffusion model as it is computationally efficient compared to models like DiGress \citep{vignac2023digressdiscretedenoisingdiffusion} which scales quadratically in the number of nodes. Given an input graph $G$ with adjacency matrix $A$, the forward corruption involves deleting the edges resulting in an empty graph as a terminal distribution. The reverse process predicts edges between active nodes, let
$s^t_i = \mathbf{1}[d^{t-1}_i \neq d^t_i]$ be a latent variable marking whether node $i$'s degree
changed at step $t$. With $d^t = \deg(A^t)$, the forward process is given by
\begin{equation}
  q\bigl(A^{1:T}, s^{1:T} \mid A^0\bigr)
  = \prod_{t=1}^{T} q\bigl(A^t \mid A^{t-1}\bigr)\, q\bigl(s^t \mid A^{t-1}, A^t\bigr),
  \label{eq:edge-forward}
\end{equation}
Marginalizing Equation \ref{eq:edge-forward} over intermediate steps gives the absorbing marginal, in which each present edge survives to step $t$ independently with probability $\bar{\alpha}_t$ giving rise to $q\!\left(A^t_{ij}=1 \mid A^0_{ij}=1\right) = \bar{\alpha}_t$ and $q\!\left(A^t_{ij}=1 \mid A^0_{ij}=0\right) = 0$
where $\bar{\alpha}_t$ follows a cosine schedule with $\bar{\alpha}_T = 0$, so that edges are only ever removed and the terminal state $A^T$ is the empty graph. The reverse process is given by
\begin{equation}
  p_\theta\bigl(A^{0:T}, s^{1:T}\bigr)
  = p\bigl(A^T\bigr) \prod_{t=1}^{T}
  p_\theta\bigl(A^{t-1} \mid A^t, s^t\bigr)\, p_\theta\bigl(s^t \mid A^t\bigr),
  \label{eq:edge-reverse}
\end{equation}
where $s^t$ in \eqref{eq:edge-forward} is a deterministic function of
$A^{t-1}, A^t$, and in \eqref{eq:edge-reverse} the model first predicts which
nodes are active and then predicts edges only between them, copying every
other pair forward unchanged. The posterior $q(s^t \mid d^t, d^0)$ has closed
form in terms of the degree and depends on $A^0$ only through
$d^0$.
\paragraph{Pre-mature generalization.} To demonstrate that existing sample level metrics are blind to the memorization, we plot the normalized mean average error (NMAE) of clustering coefficient and the WL kernel similarity for graphs generated by the diffusion model for SBMs with $n = \{64,128,256,512\}$ in Figure \ref{fig:blindmetrics}. We can observe that as the training progresses, the NMAE improves consistently (lower is better), indicating the generated graphs are 
generating graphs with the right community structure
and the WL kernel similarity is flat across the training epochs implying, the generated graphs are novel and diverse. Seeing these results one would conclude rather pre-maturely that the model is indeed generalizing. However, as we show in the next section, the model is actually memorizing the training graph, this is indicated as a growing gray band in the plot, interestingly looking at the generated samples, we can observe that the community structure of the graph only appears well inside this memorization regime and before this onset, the generated samples are structureless blobs.

\section{Conditional memorization in the reverse diffusion trajectories}
\begin{wrapfigure}{L}{0.39\textwidth}
    \centering
    \vspace{-0.5cm}
    \includegraphics[width=0.39\textwidth]{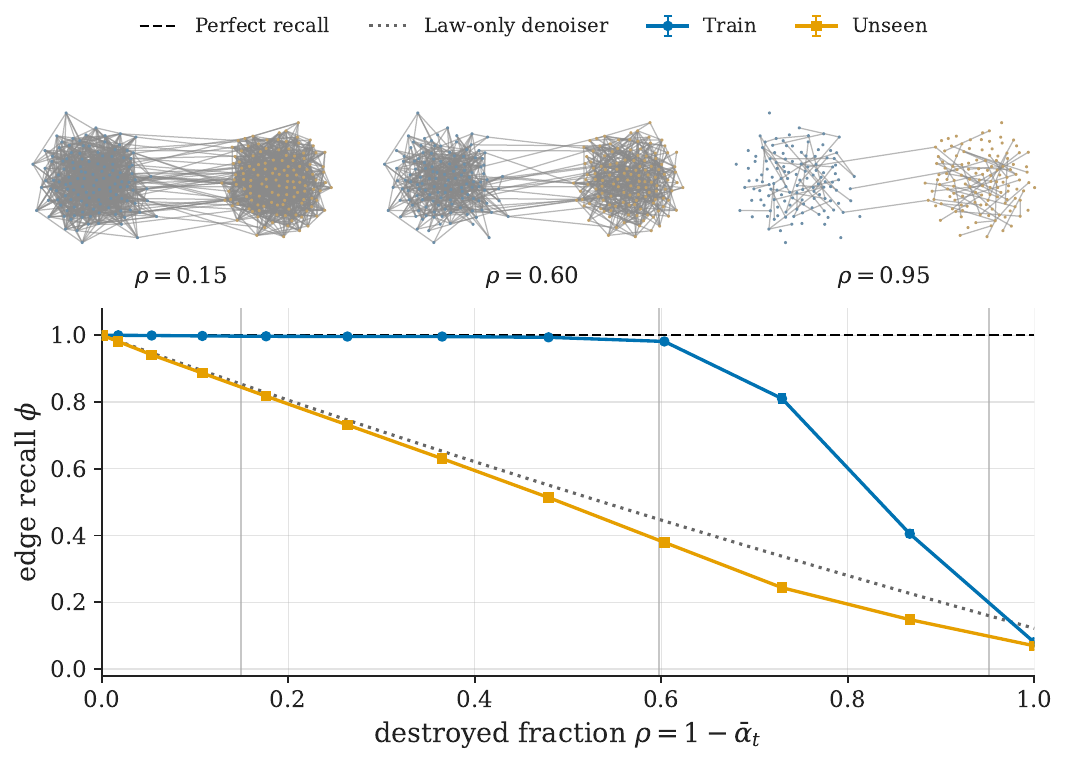}
    \caption{We measure the ability of the denoiser to recover the training graph vs. test graph and find the model exhibits a training-data bias where it can recover the training graph perfectly.}
    \vspace{-0.5cm}
    \label{fig:n256edgerecall}
\end{wrapfigure}
To show that the graph diffusion models indeed exhibit conditional memorization, we adopt the \emph{U-turn} experimental setup
proposed in \citep{garnierbrun2026biasedgeneralizationdiffusionmodels}which involves analyzing the reverse diffusion trajectories seeded with the training data vs. the test data.

\paragraph{Reference law-only denoiser.} We begin by defining a law-only denoiser that can serve as an unbiased reference. The reference denoiser knows the generating law and the corruption level, but does not have access to the graph itself. Concretely, it knows the block assignment $z$ and the block probabilities $p_{z_i z_j}$. This gives us a baseline behavior we can expect from a model that is not memorizing a particular training graph. The details of which are present in Appendix \ref{uturnapp}.

\paragraph{U-turn probe for edge recall.}
Let $A^0$ be the edge states of a reference graph, with edge set $\mathcal{E}(A^0)$ and degree sequence $d^0 = \deg(A^0)$. We corrupt $A^0$ up to step $t$ using the forward process: each edge is deleted independently with probability $\rho = 1 - \bar{\alpha}_t$, which we refer to as the destroyed fraction. We then run the reverse process from $A^t$ back to $t = 0$ to obtain a reconstruction $\tilde{A}^0$. At each reverse step the model first predicts the active nodes $s^t$ and then predicts edges only between active pairs, with $d^0$ supplied as an auxiliary feature. We measure edge recall as
\begin{equation} \label{eq:edgerecall}
  \phi_{\texttt{edge-recall}}
  = \mathbb{E}\!\left[
      \frac{\left|\mathcal{E}(\tilde{A}^0) \cap \mathcal{E}(A^0)\right|}
           {\left|\mathcal{E}(A^0)\right|}
    \right],
\end{equation}
estimated as the mean over $R$ independent reverse trajectories. More details are given in Appendix \ref{uturnapp}.
Figure~\ref{fig:n256edgerecall} plots $\phi_{\texttt{edge-recall}}$ against $\rho$ for
$n = 256$, where $\phi = 1$ denotes perfect recall, every corrupted edge restored to its
original position. The denoiser exhibits a train-data bias, for instance, for $n=256$ at $\rho = 0.60$ it restores $96.9\%$ of the corrupted edges of the training graph, against $7.7\%$ for a
denoiser that has learned only the generating law, and recall on an unseen test graph stays close to that reference throughout. We find this behavior across node sizes although the edge recall rate varies for different node sizes.

\begin{wrapfigure}{L}{0.35\textwidth}
    \centering
    \vspace{-0.5cm}
    \includegraphics[width=0.35\textwidth]{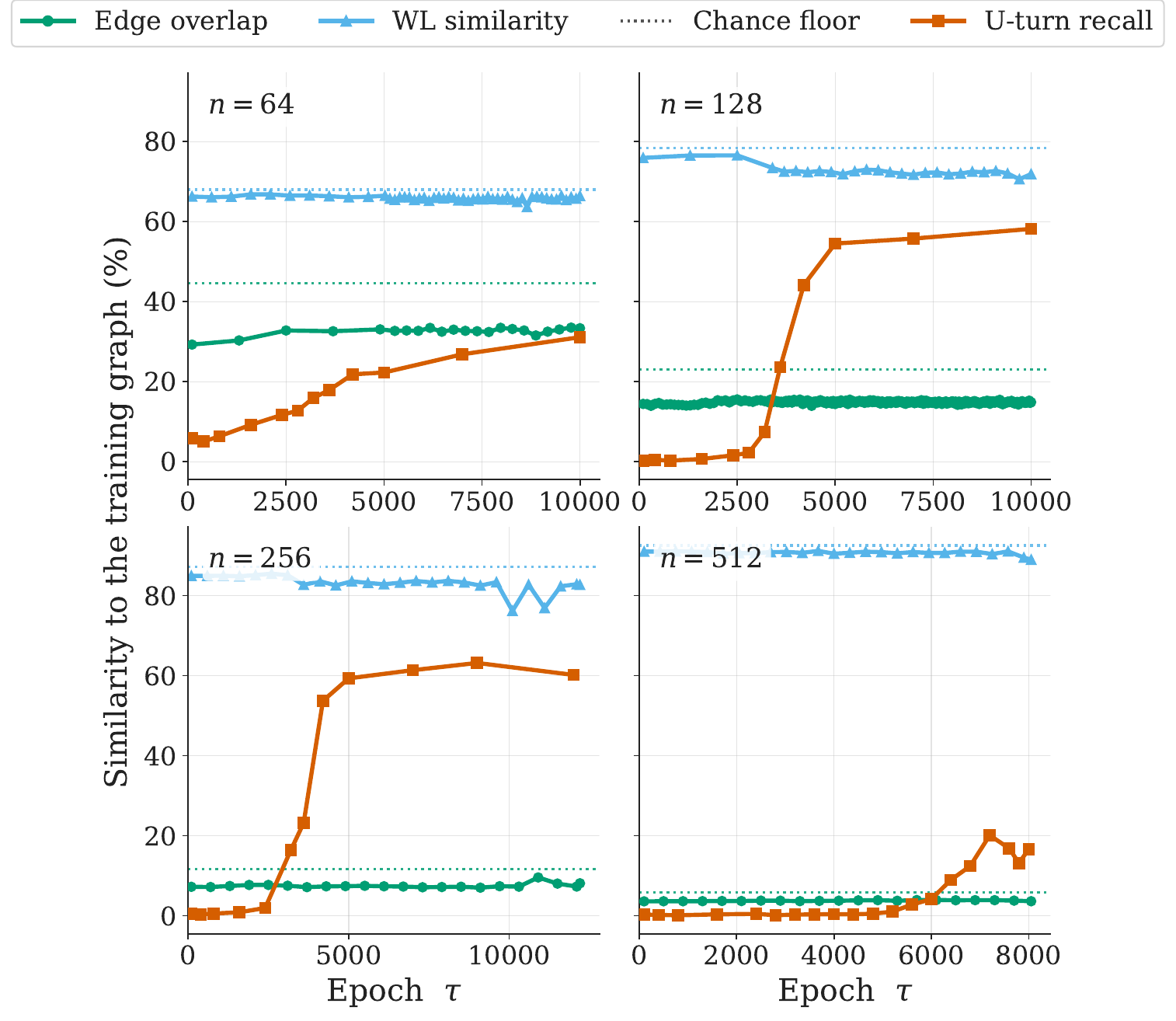}
    \caption{Analyzing the edge recall through a U-turn probe allows us to detect memorization.}
    \vspace{-0.5cm}
    \label{fig:overlapvsrecall}
\end{wrapfigure}
We term this \emph{conditional memorization}: the training instance is recoverable from the model's reverse dynamics, but only when those dynamics are initialized inside the region
of noisy states the model was trained on, a region unconditional sampling does not enter. A similar observation has been noted in contemporaneous
work \citep{kaiser2026diffusionmodelsmemorizetraining}, although our analysis suggests a different conclusion, that is, it is not that the models memorize during training and generalize in inference, it is that the sample-level metrics commonly used to assess generalization fail to capture this bias enabling a premature conclusion of the model's generalization abilities which can have unintended consequences for privacy-preserving deployment
\citep{carlini2023extractingtrainingdatadiffusion,somepalli2022diffusionartdigitalforgery}. 

\paragraph{Edge recall for memorization.} In Figure \ref{fig:overlapvsrecall}, we plot edge overlap, WL kernel similarity and edge recall from the U-turn experiment across node sizes. Edge overlap and WL similarity stay flat, which would indicate no memorization. This is an artifact of how sampling works: inference starts from an empty graph and never passes near the training graph, so any sample-level comparison is clean by construction. The U-turn probe queries the model where it was actually trained, and there edge recall climbs steadily, exposing a bias that sample-level metrics cannot see.

\paragraph{Community structure emerges only after memorization}
\citet{garnierbrun2026biasedgeneralizationdiffusionmodels} study a phenomenon of biased generalization where the test loss improves while the diffusion model continues to show unusually high ability to recover the training samples. In contrast, we observe that a graph diffusion model progressively overfits the
denoising objective, resulting in a wide generalization gap. We measure the community structure of the generated samples by their average clustering coefficient $c$, normalized against two references: the training graph's own clustering $c_{\text{train}}$,
and a floor $c_{\text{ER}}(\tau) = \bar{d}_{\text{gen}}(\tau)/(n-1)$, the clustering expected of an Erd\H{o}s--R\'enyi graph of the same mean degree, giving
$\mathcal{S}(\tau) = \frac{c_{\text{gen}}(\tau) - c_{\text{ER}}(\tau)}{c_{\text{train}} - c_{\text{ER}}(\tau)}$.
\begin{wrapfigure}{L}{0.35\textwidth}
    \centering
    \vspace{-0.5cm}
    \includegraphics[width=0.35\textwidth]{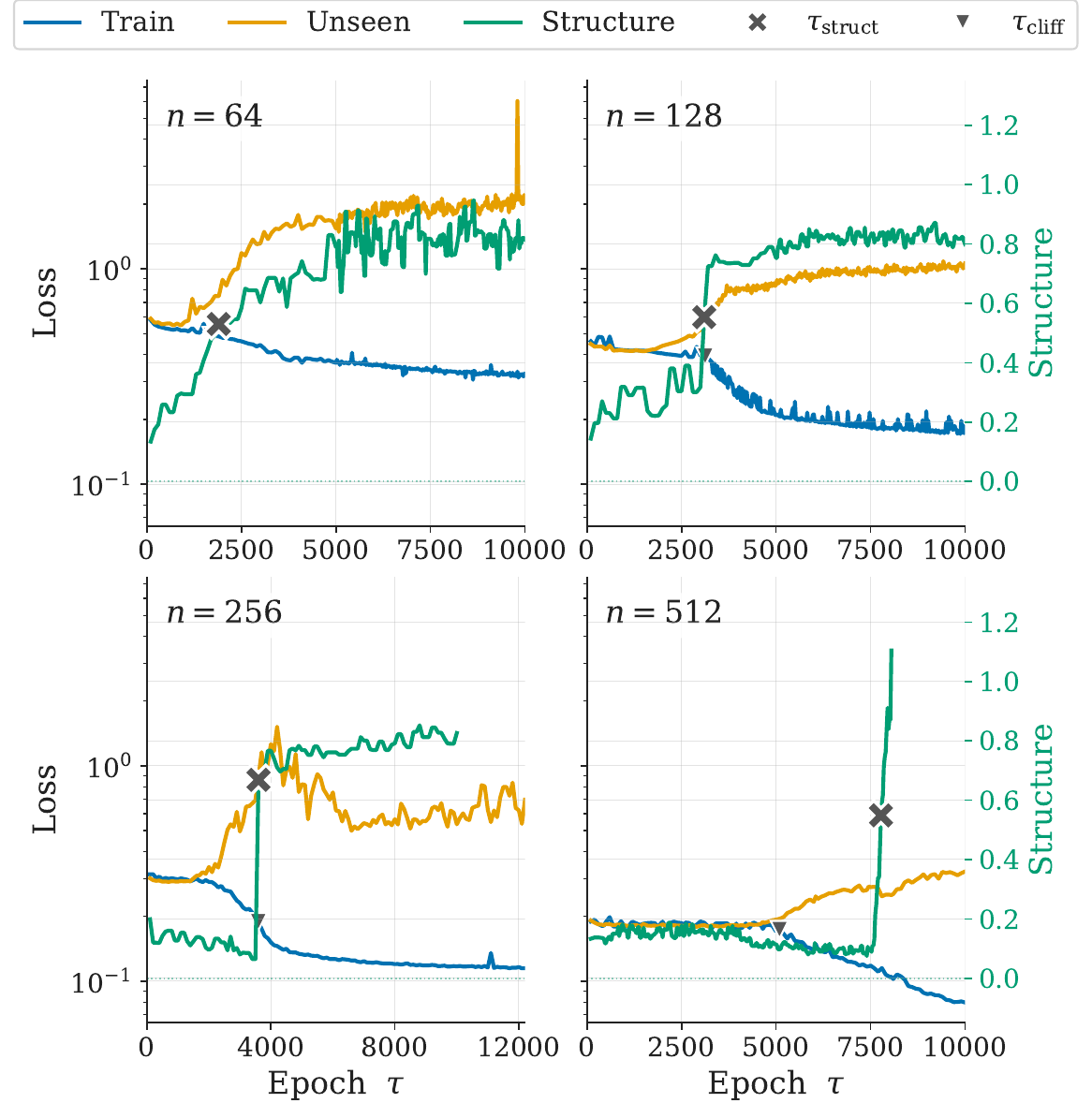}
    \caption{Train and test loss with the structure fraction $\mathcal{S}(\tau)$ across node sizes. Community structure emerges at $\tau_{\texttt{struct}}$ well after the onset of overfitting at $\tau_{\texttt{cliff}}$.}
    \vspace{-0.5cm}
    \label{fig:phasetransition}
\end{wrapfigure}
A value of $0$ means the samples are as unstructured as a random graph of their own density and $1$ means they match the training graph.\footnote{the floor is recomputed
at every evaluation from the samples' \emph{own} mean degree, because clustering rises with density and a fixed floor would credit a denser sample with structure it does not have.} We define $\tau_{\texttt{struct}}$ as the first epoch at which $\mathcal{S}(\tau)$ reaches $0.5$ and stays there for two consecutive evaluations. 

Figure~\ref{fig:phasetransition} plots $\mathcal{L}_{\texttt{train}}$ and $\mathcal{L}_{\texttt{test}}$ alongside $\mathcal{S}(\tau)$, with $\tau_{\texttt{struct}}$ annotated: community structure appears well inside the regime in which the model has already begun overfitting the training graph. The memorized regime cannot be avoided by stopping early. Table~\ref{tab:onsets} gives
the onset of instance recall $\tau_{\texttt{mem}}$ and the emergence of community
structure $\tau_{\texttt{struct}}$ at every size, and the ordering is the same in all the four cases: $\tau_{\texttt{mem}} < \tau_{\texttt{struct}}$. Every checkpoint that generates graphs with the right structure is already past the onset of instance recall, and every checkpoint before it produces structureless samples. This also makes the test loss a poor selection criterion, since its minimum returns a model that has not yet learned the community structure at all.

\section{Conclusion}
In this work we show that single-graph diffusion models exhibit what we term \emph{conditional memorization}, where the denoiser is capable of recovering the training graph with perfect recall compared to a test graph. We demonstrate this phenomenon on synthetic graphs drawn from a SBM across different node sizes. Further, we show that the community structure in the graph only appears after the onset of memorization and thus choosing a model based on test loss will result in a sub-optimal denoiser that outputs blob-like graphs.

\newpage
\medskip
\bibliographystyle{plainnat}
\bibliography{main}

\newpage
\appendix

\section{Acknowledgements}
This work was funded by the European Union ERC-2024-
STG-101163069 MALAGA.

\section{Limitations} Our experiments use a single two-block SBM, and it remains to be seen whether the same picture holds for other graph families where the structure that emerges is less cleanly summarized by a single statistic. Studying if the same trend is exhibited by other diffusion models like \citep{vignac2023digressdiscretedenoisingdiffusion} is also a future direction. The dependence on node size is also only partially probed: our four sizes suggest a trend, but we do not characterize how $\tau_\texttt{cliff}$ and $\tau_\texttt{struct}$ scale with node size. Finally, we train on a single graph throughout. How the bias measured by the U-turn probe behaves as the training set grows beyond one graph is left for future work.

\section{Related work}
\paragraph{Generalization vs. Memorization.} Understanding why diffusion models \citep{ddpm,d3pm} generalize is interesting both theoretically and from a privacy perspective \citep{somepalli2022diffusionartdigitalforgery,carlini2023extractingtrainingdatadiffusion}. Many recent works have explored a variety of avenues to explain the generalization vs memorization dichotomy in diffusion models. \citet{kadkhodaie2024generalization} attribute the generalization ability of the diffusion models to the inductive bias while other works explore implicit regularization and network capacity \citep{bonnaire2025why,Biroli_2024,ambrogioni2024statisticalthermodynamicsgenerativediffusion,farghly2025implicitregularisationdiffusionmodels,gu2025memorizationdiffusionmodels} as possible arguments. \citet{garnierbrun2026biasedgeneralizationdiffusionmodels} show that diffusion models exhibit a biased generalization, where even though the test loss improves, the model shows unusually high affinity to generate some of the training samples. Similarly, \citet{kaiser2026diffusionmodelsmemorizetraining} show that models that generalize well at the sample level overfit the denoising objective resulting in a generalization gap between the train samples and test samples.
\paragraph{Memorization in graph diffusion models.} Graph diffusion models have received comparatively little attention along these lines. Existing works on discrete graph diffusion and flow matching \citep{vignac2023digressdiscretedenoisingdiffusion,qin2025sparsediff,chen2023efficientdegreeguidedgraphgeneration,osman2025lgdclatentgraphdiffusion,jiang2026bureswasserstein} target generation quality and efficient scaling to large graphs. The notion of memorization is ill-defined for GDMs, since checking whether a model has memorized a graph amounts to solving the graph isomorphism problem as each training graph has $\Omega(n!)$ graphs in its isomorphism class. This is further worsened by the message-passing GNN architectures used as denoisers, which are permutation equivariant by construction and so invariant to node ordering. Works such as \citep{laabid2025equivariant,yan2024swingnnrethinkingpermutationinvariance,lawrence2025improvingequivariantnetworksprobabilistic} show graph diffusion models cannot exactly copy their training graphs. In the single-graph training paradigm such as EDGE \citep{chen2023efficientdegreeguidedgraphgeneration}, edge overlap \citep{edgedependency,powerindependent} is used to assess whether generated graphs are highly similar to the reference training graph. Inspired by the notion of biased generalization in image diffusion models \citep{garnierbrun2026biasedgeneralizationdiffusionmodels,kaiser2026diffusionmodelsmemorizetraining}, we study the learning dynamics of single-graph diffusion models, and find that they exhibit the same signature — a widening train–test generalization gap and a biased reverse process that leads to a behavior we call \emph{conditional memorization} where the model has the ability to recover the training graph exactly compared to a test graph. Further, our analysis reveals that the community structure only appears after the onset of this memorization which early stopping cannot mitigate.

\section{Details on U-turn experiment} \label{uturnapp}
We adopt the experimental setup of \citep{garnierbrun2026biasedgeneralizationdiffusionmodels} to demonstrate train-data bias. Given a fixed checkpoint and a graph $G$ on $n$ nodes with edge set $E(G)$, for a chosen timestep $t$ we apply the model's own
forward kernel to $G$'s clean state $G_0$, obtaining a corrupted state $G_t$,
and then run the model's reverse process from $G_t$ down to $G_0$,
generating a reconstruction $\tilde G$. We measure edge recall
\begin{equation} \label{edgerecalleqn}
  \phi_t(G) \;=\; \frac{|E(\tilde G)\cap E(G)|}{|E(G)|}  
\end{equation}

We report $t$ on the interpretable axis $\rho = 1-\bar\alpha_t$, the expected fraction of $G$'s edges removed by step $t$. All runs use $T = 256$ diffusion steps with a cosine schedule. We evaluate at
$12$ timesteps evenly spaced over the full range,
$t \in \{5, 28, 51, 74, 96, 119, 142, 165, 188, 210, 233, 255\}$, covering
$\rho$ from $0.002$ to $1$. 

At each (checkpoint, $t$) we run $R$ independent U-turns of the training graph,
differing only in the forward and reverse randomness, and report the mean and
the standard error across them. We repeat the identical procedure on fresh graphs drawn from the same stochastic block model never seen in training.

\paragraph{Reference law-only denoiser.} First, we begin by defining a law-only denoiser that can serve as an unbiased reference. The reference denoiser knows the generating law and the corruption level, but does not have access to the graph itself. Concretely, it knows the block assignment $z$ and the block probabilities $p_{z_i z_j}$. The posterior factorizes as $p(A^0 \mid A^t, z) = \prod_{i<j} p(A^0_{ij} \mid A^t_{ij}, z)$. The forward process only deletes edges, so an observed
present pair must have been an edge, $p(A^0_{ij}=1 \mid A^t_{ij}=1) = 1$, while for an absent pair Bayes' rule gives
\begin{equation}
  \beta(p,\rho)
  \;=\; p\!\left(A^0_{ij}=1 \mid A^t_{ij}=0\right)
  \;=\; \frac{\rho\, p}{\rho\, p + (1-p)},
\end{equation}

In the reverse process the edges are sampled independently with probability $
\beta(p,\rho)$, thus the law-only denoiser is capable of restoring the correct \emph{number} of edges in expectation, but cannot recall the exact pairs. Averaging over true edges, a given edge either survives corruption or is deleted and predicted, giving
\begin{equation}
\phi_{\mathrm{law}}(\rho)
  \;=\; \underbrace{(1-\rho)}_{\text{survived}}
      \;+\; \underbrace{\rho\,\beta(p,\rho)}_{\text{restored}}
  \;=\; (1-\rho) + \frac{p\,\rho^{2}}{p\,\rho + 1 - p}
\end{equation}

\begin{figure}[h]
    \centering
    \begin{subfigure}{0.49\linewidth}
        \centering
        \includegraphics[width=\linewidth]{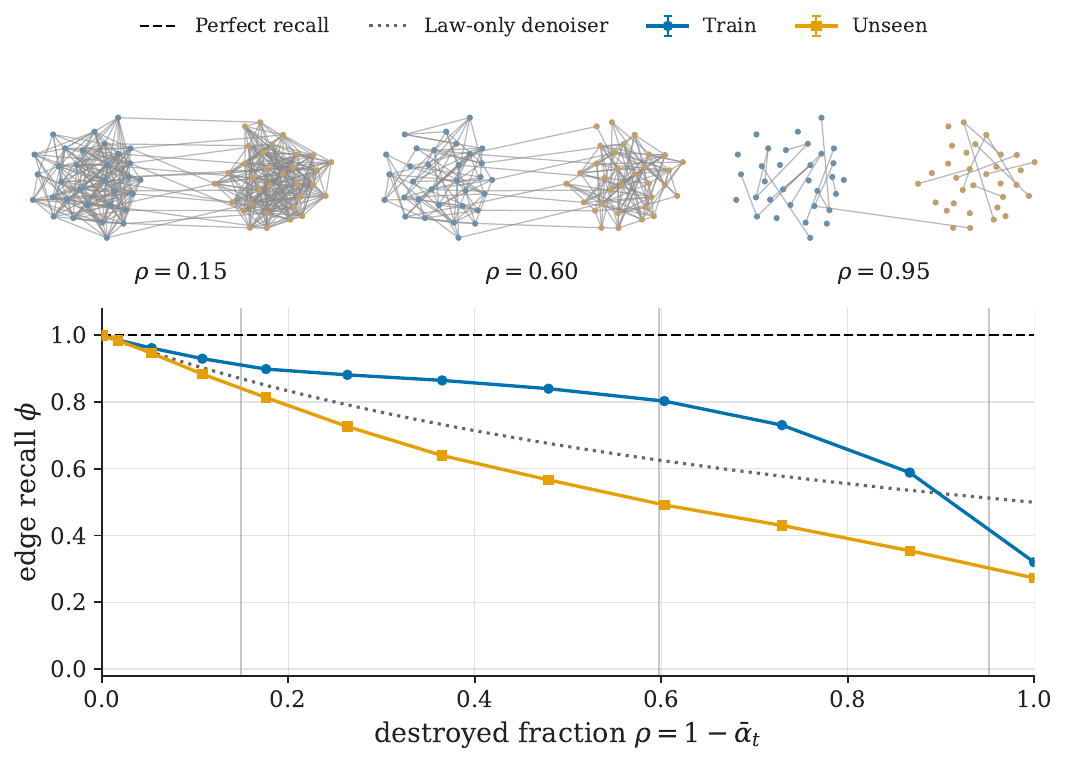}
        \caption{n64.}
          \label{fig:n64}
    \end{subfigure}
    \begin{subfigure}{0.49\linewidth}
        \centering
        \includegraphics[width=\linewidth]{images/sbm_n256_k2_m7p0_seed0_runcosine256steps_ep11999_toy_model.pdf}
        \caption{n256.}
          \label{fig:n256}
    \end{subfigure}
    \begin{subfigure}{0.49\linewidth}
        \centering
        \includegraphics[width=\linewidth]{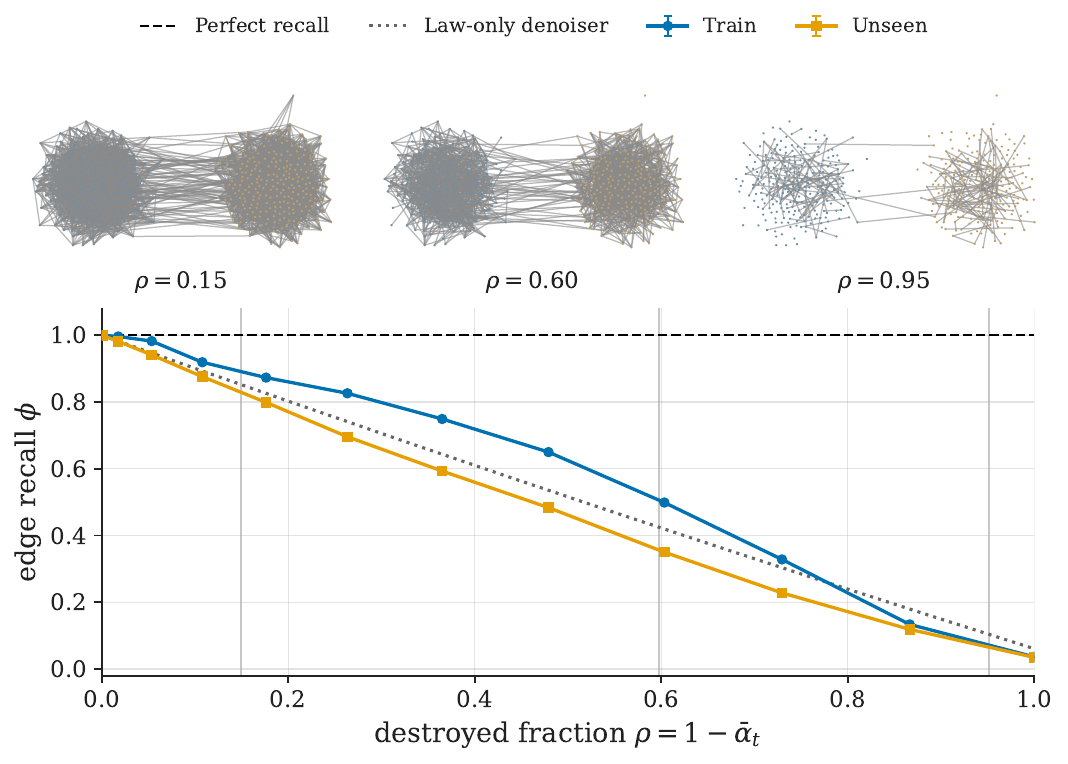}
        \caption{n512.}
        \label{fig:n512}
    \end{subfigure}
    \caption{Edge recall across node sizes.}
    \label{fig:edgerecallacross}
\end{figure}

\paragraph{Checkpoint admissibility.} Late in training at $n = 512$ the sampler develops a density instability in
which a growing fraction of reverse trajectories emit far too many edges;
recall is inflated one-sidedly under this failure, since a denser
reconstruction intersects the target more often. We therefore discard any
checkpoint whose reconstructions exceed $1.15\times$ the training graph's edge
density on \emph{either} arm, evaluated at the $\rho$ of peak separation. This
removes $5$ of $25$ checkpoints at $n = 512$ (epochs $8399$--$9999$, densities
$1.75$--$2.26\times$) and none at any other size. All reported $n = 512$
quantities are therefore truncated at epoch $8050$. In Figures \ref{fig:edgerecallacross}, we show the edge recall from the U-turn experiment for different node sizes.

\paragraph{Early stopping cannot mitigate memorization.}Memorization begins before the model produces structured samples, at every
size. $\tau_{\mathrm{mem}}$ is the first checkpoint at which the peak recall
$\max_\rho \Delta\phi_{\texttt{edge-recall}}$ exceeds a common absolute floor of
$0.02$; $\tau_{\mathrm{struct}}$ is the first checkpoint at which the structure
fraction of \emph{unconditionally} generated samples crosses the E-R floor;
$\tau_{\mathrm{gen}}$ minimises the per-node-pair loss on held-out graphs from
the same law. At $n=64$ the recall advantage is already above the floor at the
earliest available checkpoint, so $\tau_{\mathrm{mem}}$ is bounded rather than
measured. At $n=512$ the held-out loss has no interior minimum within the
scored range. The ordering
$\tau_{\mathrm{gen}} < \tau_{\mathrm{mem}} < \tau_{\mathrm{struct}}$ holds
wherever all three are defined: early stopping at the held-out optimum yields a
model that has not yet learnt the block structure, and every checkpoint that
has learnt it is already past the onset of instance recall. This is shown in Table \ref{tab:onsets}.

\begin{table}[h]
\centering
\caption{Memorization onset happens before the emergence of community structure.}
\begin{tabular}{@{}lrrrr@{}}
\toprule
 & $n=64$ & $n=128$ & $n=256$ & $n=512$ \\
\midrule
$\tau_{\mathrm{gen}}$ \; (argmin held-out loss) & 500 & 1300 & 700 & --- \\
$\tau_{\mathrm{mem}}$ \; (recall onset)         & $\leq 99$ & 2799 & 3199 & 5599 \\
$\tau_{\mathrm{struct}}$ (structure in samples) & 1900 & 3100 & 3600 & 7775 \\
\midrule
$\tau_{\mathrm{struct}} - \tau_{\mathrm{mem}}$  & $\geq\!+1801$ & $+301$ & $+401$ & $+2176$ \\
\bottomrule
\end{tabular}

\label{tab:onsets}
\end{table}

It is possible the model might artificially show high edge recall simply by predicting too many edges, to account for this, we also compute the following: for a target graph $G$ and its U-turn reconstruction $\widetilde{G}$, let
\begin{equation}
 M = |E(G)|,\qquad
K = |E(\widetilde{G})|,\qquad
I = |E(G)\cap E(\widetilde{G})|.   
\end{equation}
The edge recall and edge F1 score for one reverse trajectory are
\begin{equation}
R(G,\widetilde{G})
=
\frac{I}{M}
=
\frac{|E(G)\cap E(\widetilde{G})|}{|E(G)|},  
\end{equation}
and
\begin{equation}
 F_1(G,\widetilde{G})
=
\frac{2I}{M+K}
=
\frac{2|E(G)\cap E(\widetilde{G})|}
{|E(G)|+|E(\widetilde{G})|}.   
\end{equation}

For checkpoint $\tau$ and corruption level $t$, we average each metric over
independent reverse trajectories. We then define the train--unseen gaps as
\[
\Delta R_{\tau,t}
=
\overline{R}_{\tau,t}(G_{\mathrm{train}})
-
\frac{1}{J}
\sum_{j=1}^{J}
\overline{R}_{\tau,t}(G^{\mathrm{unseen}}_j),
\]
and
\[
\Delta F_{1,\tau,t}
=
\overline{F}_{1,\tau,t}(G_{\mathrm{train}})
-
\frac{1}{J}
\sum_{j=1}^{J}
\overline{F}_{1,\tau,t}(G^{\mathrm{unseen}}_j),
\]
where $J$ is the number of unseen graphs and the overline denotes an average
over reverse trajectories.

Finally, we report the maximum gaps over the corruption grid $\mathcal{T}$:
\[
\Delta R_{\max}(\tau)
=
\max_{t\in\mathcal{T}}
\Delta R_{\tau,t},
\qquad
\Delta F_{1,\max}(\tau)
=
\max_{t\in\mathcal{T}}
\Delta F_{1,\tau,t}.
\]
This is shown in Table \ref{tab:f1tab}.
\begin{table}[h]
\centering
\caption{}
\label{tab:f1tab}
\resizebox{5cm}{!}{%
\begin{tabular}{@{}ccc@{}}
\toprule
n &
  $\Delta_{\texttt{max}} \texttt{Recall}$ &
  $\Delta_{\texttt{max}} \texttt{F1}$ \\ \midrule
64  & 0.30±0.02 & 0.30±0.02 \\
128 & 0.58±0.01 & 0.58±0.01 \\
256 & 0.62±0.01 & 0.60±0.00 \\
512 & 0.48±0.01 & 0.47±0.00 \\ \bottomrule
\end{tabular}%
}
\end{table}

\section{Experimental setup} \label{exptsetup}

\subsection{Data generation}
In Table \ref{tab:sbmparams} we show the hyperparameters used to generate our SBM training graphs. We choose the simplest two-block setting as this allows us to control the underlying generation process and it also provides the advantage of having an interpretation of what emergence of structure means.
\begin{table}[h]
\centering
\caption{SBM parameters for data generation.}
\label{tab:sbmparams}
\resizebox{7cm}{!}{%
\begin{tabular}{@{}ccccc@{}}
\toprule
n & $p_{in}$ & $p_{out}$ & \begin{tabular}[c]{@{}c@{}}Mean \\ degree\end{tabular} & \begin{tabular}[c]{@{}c@{}}Edges\\ (intra/inter)\end{tabular} \\ \midrule
64  & 0.4995 & 0.0161 & 17.00 & 544 (519/25)    \\
128 & 0.2458 & 0.0081 & 16.19 & 1036 (1000/36)  \\
256 & 0.1219 & 0.0040 & 16.43 & 2103 (2045/58)  \\
512 & 0.0607 & 0.0020 & 16.12 & 4126 (4014/112) \\ \bottomrule
\end{tabular}%
}
\end{table}

\subsection{Computing metrics.}
In Figure \ref{fig:overlapvsrecall} we plot the chance floor for edge overlap and WL kernel similarity metrics, this allows us to get a null behavior to compare against since neither marginal statistic is zero for an unmemorized model, so each is reported against its own measured floor.
\paragraph{Edge overlap.} For labelled graphs $G,H$ on a common vertex set we write $\mathrm{ov}(G,H)=|E(G)\cap E(H)|$, reported as a percentage of
$|E(G_{\mathrm{train}})|$. The floor is the overlap that two \emph{unrelated}
graphs from the same law already share: we draw $K=16$ independent graphs
$F_1,\dots,F_K$ from the same stochastic block model as $G_{\mathrm{train}}$
(disjoint seeds) and take the mean over all $\binom{K}{2}=120$ pairs,
\[
  \mathrm{ov}_{\mathrm{chance}}
  \;=\;
  \binom{K}{2}^{-1}\sum_{i<j}\mathrm{ov}(F_i,F_j).
\]
This quantity is model-independent, so it is computed once and reused at every
checkpoint. At each checkpoint we draw $32$ samples and report
$\overline{\mathrm{ov}}(G_{\mathrm{gen}},G_{\mathrm{train}})$ against it.

\paragraph{WL kernel similarity.} We use the normalized WL subtree kernel
with $h=3$ refinement rounds. Because the model has no node features, the
initial color of every vertex is its degree; colors are refined by
$c^{(r+1)}(v)=\big(c^{(r)}(v),\,\{\!\{c^{(r)}(u):u\in N(v)\}\!\}\big)$, and
$\mathrm{WL}(G,H)$ is the cosine similarity of the concatenated color
histograms over rounds $0,\dots,h$, so that $\mathrm{WL}=1$ exactly for a
relabeled copy. The floor is a \emph{degree-preserving} null: we generate $8$
rewirings $R_1,\dots,R_8$ of $G_{\mathrm{train}}$ by double-edge swaps
($10\,|E|$ attempted swaps, each accepted only if it creates no self-loop or
multi-edge), which preserve every vertex degree exactly while destroying the
block structure. Two quantities are reported,
\[
  \mathrm{WL}_{\mathrm{floor}}
  = \tfrac{1}{8}\sum_{k}\mathrm{WL}(G_{\mathrm{train}},R_k),
  \qquad
  \mathrm{WL}_{\mathrm{excess}}
  = \overline{\mathrm{WL}}(G_{\mathrm{gen}},G_{\mathrm{train}})
  - \overline{\mathrm{WL}}(G_{\mathrm{gen}},R_{\cdot}),
\]
the second being the part of the similarity that is specific to the training
\emph{instance} rather than to its degree sequence.

\subsection{Training details}
We use the experimental setup of EDGE \citep{chen2023efficientdegreeguidedgraphgeneration} as this is more computationally efficient compared to models like DiGress \citep{vignac2023digressdiscretedenoisingdiffusion} which scale quadratically with node size. Across all node sizes we train the models to 10K epochs. We train on a single graph. At each step we sample a timestep $t$, which sets the corruption level, apply the forward noising process at that level, and train the model to denoise the result back to the clean graph. Since there is only one graph, the same graph is presented many times during training; what changes from step to step is the timestep drawn and the particular corruption realized. Held-out loss is measured on fresh graphs drawn from the same block model, which the model never sees during training.

\begin{table}[h]
\centering
\caption{Training hyperparameters for the single-graph experiments. All
settings are shared across node sizes except where noted.}
\label{tab:single-graph-hyperparams}
\begin{tabular}{ll}
\toprule
Parameter & Value \\
\midrule
\multicolumn{2}{l}{\emph{Diffusion}} \\
Diffusion steps $T$            & 256 \\
Noise schedule                 & Cosine \\
Forward process                & Absorbing (edge deletion) \\
\midrule
\multicolumn{2}{l}{\emph{Architecture}} \\
Backbone                       & Degree-guided transformer GNN \\
Hidden dimension               & 128 \\
Attention heads (per layer)    & 8, 8, 8, 8, 1 \\
Node feature                   & Degree \\
Max degree                     & 26 / 27 / 27 / 30 \\
\midrule
\multicolumn{2}{l}{\emph{Optimisation}} \\
Optimiser                      & Adam \\
Learning rate                  & $1\times10^{-4}$ \\
Gradient clipping              & 1.0 \\
Warmup                         & None \\
Batch size                     & 32 \\
Steps per epoch                & 256 \\
Epochs                         & 10{,}000 (12{,}200 for $n=256$) \\
Seed                           & 0 \\
\bottomrule
\end{tabular}
\end{table}

\begin{figure}[h]
\centering
    \begin{subfigure}{0.46\linewidth}
        \centering
        \includegraphics[width=\linewidth]{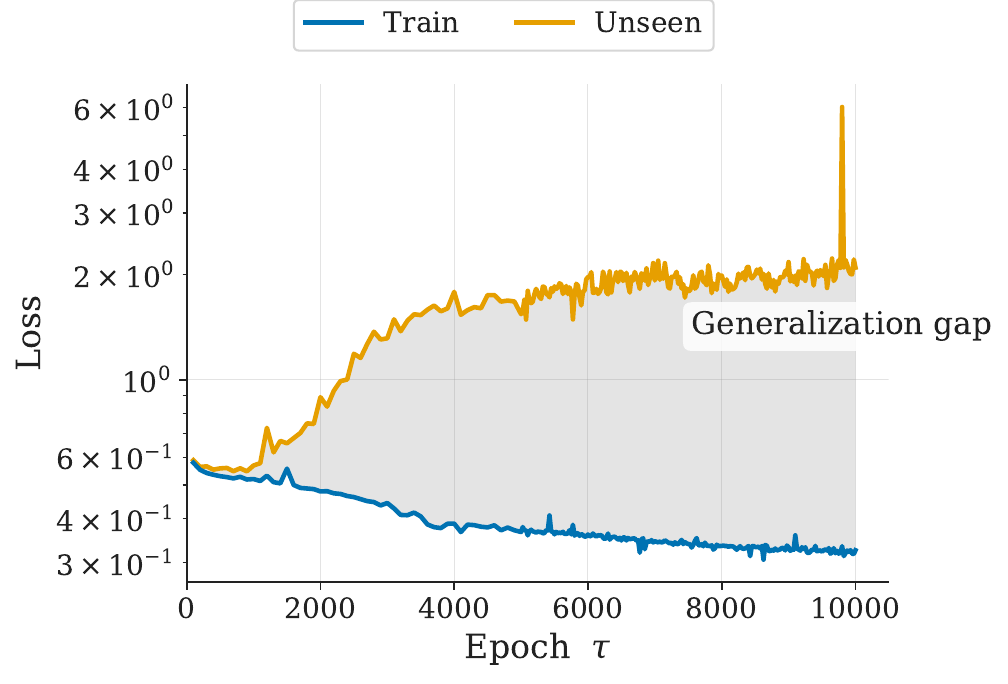}
        \caption{Generalization gap for $n = 64$.}
          \label{fig:n64}
    \end{subfigure}
    \centering
    \begin{subfigure}{0.46\linewidth}
        \centering
        \includegraphics[width=\linewidth]{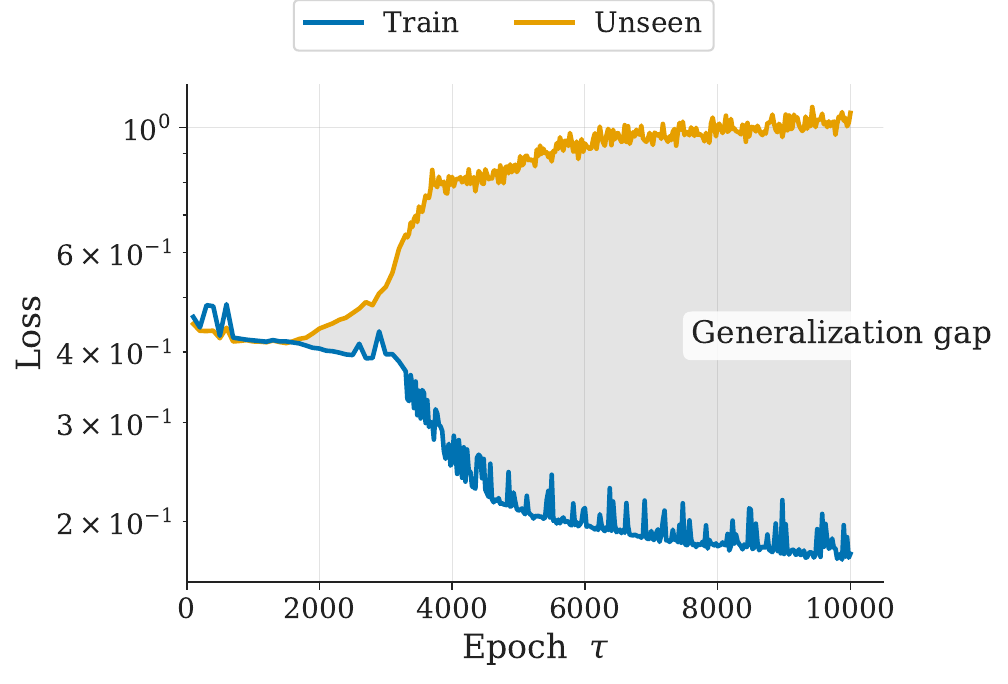}
        \caption{Generalization gap for $n = 128$.}
          \label{fig:n64}
    \end{subfigure}
    \begin{subfigure}{0.46\linewidth}
        \centering
        \includegraphics[width=\linewidth]{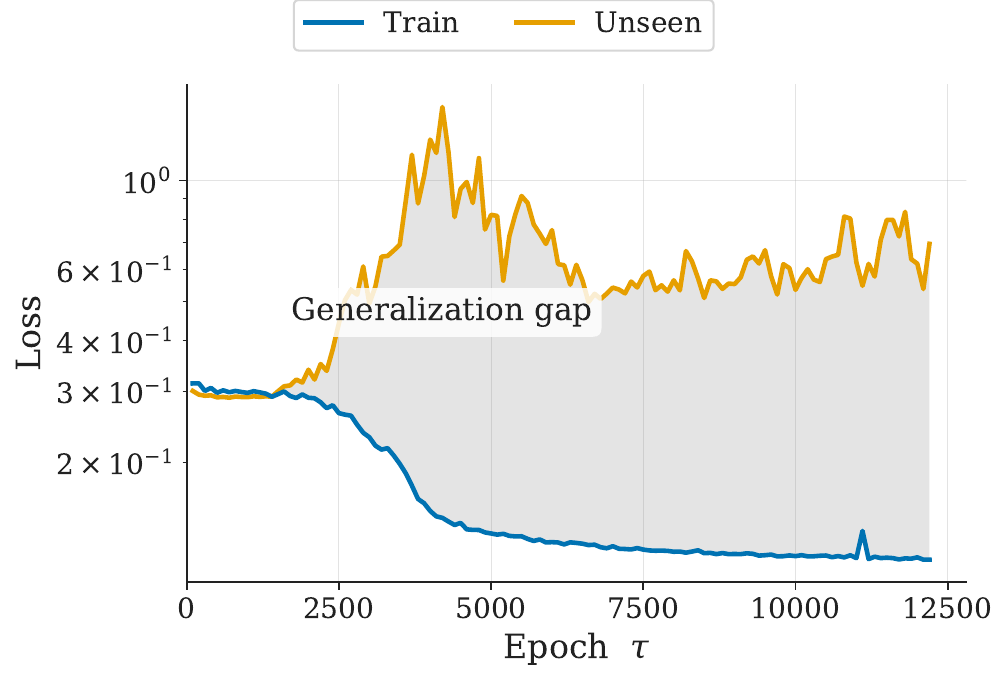}
        \caption{Generalization gap for $n=256$.}
        \label{fig:n256}   
    \end{subfigure}
    \begin{subfigure}{0.46\linewidth}
        \centering
        \includegraphics[width=\linewidth]{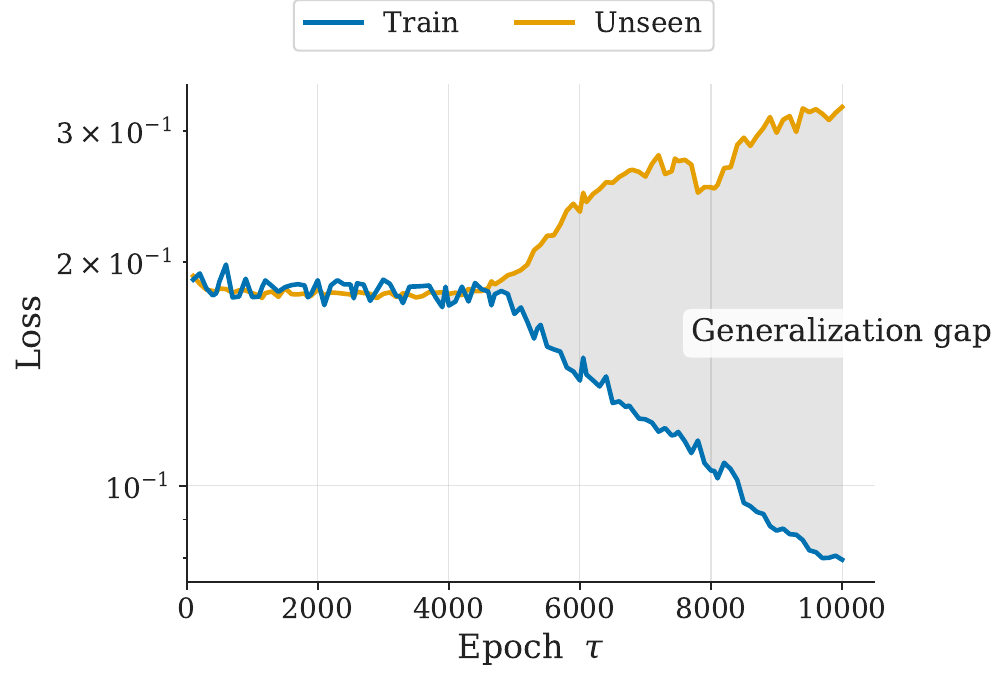}
        \caption{Generalization gap for $n=512$.}
        \label{fig:n512}
    \end{subfigure}
    \caption{Generalization gap across node sizes.}
    \label{fig:losswindowsothers}
\end{figure}

\paragraph{Loss windows.} In Figures \ref{fig:losswindowsothers} we track the test loss on held-out test graphs to demonstrate the growing generalization gap. For this we generate 8 fresh graphs from the same SBM law as used for generating the training graph.

\paragraph{Structure emergence.} For Figure \ref{fig:phasetransition} at each evaluation epoch (25) we generate 64 graphs and the clustering coefficient and mean degree is computed and averaged. After computing $\mathcal{S}(\tau)$ we use a rolling median with window 3 to smooth out the values and $\tau_{\texttt{struct}}$ is defined as the first epoch where this smoothed value reaches 0.5 and stays there for 2 consecutive epochs.

\paragraph{Edge recall.} For the U-turn probe we average over $R$ independent reverse trajectories per checkpoint. The details are provided in the table below, where we use smaller R sizes for increasing node size.

\begin{table}[h]
\centering
\caption{U-turn experiment R draws.}
\label{tab:rdraws}
\resizebox{3cm}{!}{%
\begin{tabular}{@{}ccc@{}}
\toprule
n   & R  & Checkpoints \\ \midrule
64  & 50 & 12          \\
128 & 40 & 12          \\
256 & 30 & 12          \\
512 & 20 & 12          \\ \bottomrule
\end{tabular}%
}
\end{table}

\section{Loss windows}
We plot the generalization gap across node sizes.

\end{document}